\documentclass[pdflatex,sn-nature]{sn-jnl}

\usepackage{graphicx}%
\usepackage{multirow}%
\usepackage{amsmath,amssymb,amsfonts}%
\usepackage{amsthm}%
\usepackage{mathrsfs}%
\usepackage{appendix}%
\usepackage{xcolor}%
\usepackage{textcomp}%
\usepackage{manyfoot}%
\usepackage{booktabs}%
\usepackage{algorithm}%
\usepackage{algorithmicx}%
\usepackage{algpseudocode}%
\usepackage{listings}%
\usepackage{caption}

\usepackage{lineno}

\usepackage{threeparttable} 

\theoremstyle{thmstyleone}%
\theoremstyle{thmstyletwo}%

\theoremstyle{thmstylethree}%

\begin{document}

\graphicspath{/figures/}

\title{Field-Space Attention for Structure-Preserving Earth System Transformers}




\author*[1]{\fnm{Maximilian} \sur{Witte}}\email{witte@dkrz.de}

\author[1]{\fnm{Meuer} \sur{Johannes}}\email{meuer@dkrz.de}

\author[1]{\fnm{Étienne} \sur{Plésiat}}\email{plesiat@dkrz.de}

\author[1]{\fnm{Christopher} \sur{Kadow}}\email{kadow@dkrz.de}

\affil*[1]{\orgname{German Climate Computing Center}, \orgaddress{\street{Bundesstrasse 45a}, \city{Hamburg}, \postcode{20146}, \state{Hamburg}, \country{Country}}}


\abstract{Accurate and physically consistent modeling of Earth system dynamics requires machine-learning architectures that operate directly on continuous geophysical fields and preserve their underlying geometric structure. Here we introduce Field-Space attention, a mechanism for Earth system Transformers that computes attention in the physical domain rather than in a learned latent space. By maintaining all intermediate representations as continuous fields on the sphere, the architecture enables interpretable internal states and facilitates the enforcement of scientific constraints. The model employs a fixed, non-learned multiscale decomposition and learns structure-preserving deformations of the input field, allowing coherent integration of coarse and fine-scale information while avoiding the optimization instabilities characteristic of standard single-scale Vision Transformers. Applied to global temperature super-resolution on a HEALPix grid, Field-Space Transformers converge more rapidly and stably than conventional Vision Transformers and U-Net baselines, while requiring substantially fewer parameters. The explicit preservation of field structure throughout the network allows physical and statistical priors to be embedded directly into the architecture, yielding improved fidelity and reliability in data-driven Earth system modeling. These results position Field-Space Attention as a compact, interpretable, and physically grounded building block for next-generation Earth system prediction and generative modeling frameworks.}




\maketitle

\section*{Introduction}
Recent advances in deep learning have enabled state-of-the-art performance across a broad spectrum of applications involving continuous spatiotemporal data in Earth system science, computational physics, and climate and weather modeling. In particular, data-driven models have achieved strong results on core geophysical tasks such as statistical downscaling (super-resolution) \cite{Harder2023,Mardani2025}, spatiotemporal infilling \cite{Kadow2020,Meuer2025,Plesiat2025}, and short- to medium-range prediction \cite{Burgh-Day2023Machine,Chen2023Iterative,zhuang2025,Bi2025,Price2024,bonev2025,Allen2025}. Beyond task-specific architectures, foundation model approaches have emerged that learn general-purpose atmospheric representations from heterogeneous datasets \cite{lessig2023,2024zhu,BrenowitzEtAl2025,Bodnar2025}, following trends in natural-language and vision modeling. These developments reflect a broader shift towards complement, accelerating, or augmenting numerical Earth system models with machine-learning surrogates that operate directly on global, multivariate physical fields \cite{irrgang2021,Kochkov2024,stevens2024}.

Across these applications, emulating physical systems typically benefits from well-defined structural priors: coarse-resolution input fields in super-resolution tasks \cite{Witte2025}, previous time steps in prediction, or sparse observations in infilling. However, most modern neural architectures process such inputs by abstracting them into high-dimensional latent spaces. Convolution-based models impose strong inductive biases through hierarchical feature extraction and locality, but inherently limit long-range information exchange \cite{Wu2023}. Vision Transformers (ViTs) mitigate this limitation by learning patch-wise latent embeddings and propagating information globally via self-attention \cite{dosovitskiy2021}. Multi-scale ViTs extend this paradigm by constructing hierarchical latent tokens that communicate across resolutions \cite{liu2021swin,Li2024,carlsson2024}. Coarser tokens aggregate broad spatial context, while finer tokens preserve local detail, enabling information to flow efficiently between scales. This multi-resolution design improves both computational efficiency and the ability to represent long-range dependencies without sacrificing fine-scale structures \cite{liu2021swin,Li2024,carlsson2024}.\\
Despite strong empirical performance, these architectures share a fundamental limitation. The information flow is ultimately mediated through learned latent representations. This imposes constraints on expressive power, convergence, and flexibility. First, reliance on latent abstractions makes the models data-hungry, especially for architectures lacking strong inductive biases like Vision Transformers \cite{Akkaya2024}. Thus, optimization from scratch becomes difficult because early-stage latent representations are poorly defined and attention mechanisms struggle to form coherent structures without substantial training signal \cite{lessig2023}. Second, operating in a latent space restricts the enforcement of physical constraints to the network’s final output. For instance, physical constraints, conservation properties, or denoising priors can typically only be imposed post hoc, rather than throughout the internal computation. Third, learning in latent space fixes the maximum spatial resolution of the model and does not allow for a natural way to increase it after training without substantially modifying the architecture.\\
To address these limitations, we introduce \emph{Field-Space Attention}, a residual multi-scale attention mechanism that operates directly on a fixed spherical decomposition of the input field. Each layer performs a structure-preserving update in physical space, so intermediate states remain valid, globally defined fields. Building Field-Space Transformers from these blocks yields rapid and stable training from scratch and favorable parameter scaling on global near-surface temperature downscaling, outperforming Vision Transformer and Transformer-U-Net baselines at substantially smaller model sizes. Across experiments, learning structure-preserving deformations in physical space provides a compact alternative to latent-space processing for accurate, interpretable and multi-scale-consistent modeling of continuous geophysical fields.

\section*{Field-Space Attention: Learning in Physical Space}
\begin{figure}[h!]
  \centering
  \includegraphics[width=1.0\textwidth]{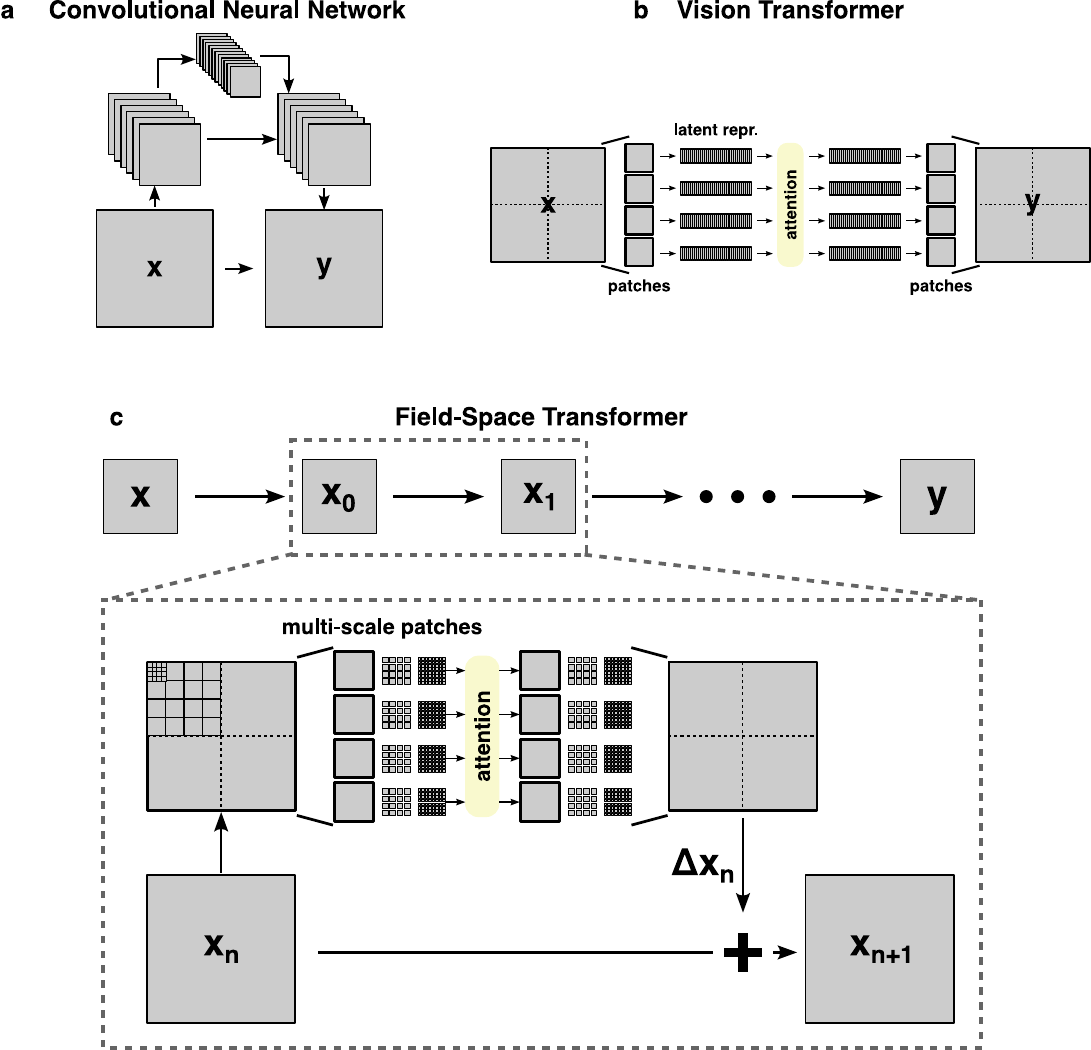}
   \caption{\textbf{Conceptual comparison of convolutional networks, Vision Transformers and the proposed Field-Space Transformer.}
  \textbf{a}, Convolutional models map the input field $x$ to the output $y$ through stacked local convolutions, typically using an encoder-decoder with skip connections.
  \textbf{b}, A Vision Transformer maps $x$ to patch tokens, transforms them by self-attention in a latent token space, and decodes tokens back to patches to obtain $y$.
  \textbf{c}, The Field-Space Transformer maintains a field-valued state: it forms multi-scale patches, applies attention across scales, and predicts a residual update $\Delta x_n$ that is added to the current field state $x_n$ to produce $x_{n+1}$.}
  \label{fig:models}
\end{figure}

\begin{figure}[ht]
  \centering
  \def\svgwidth{0.8\columnwidth}  
  \includegraphics[width=1.0\textwidth]{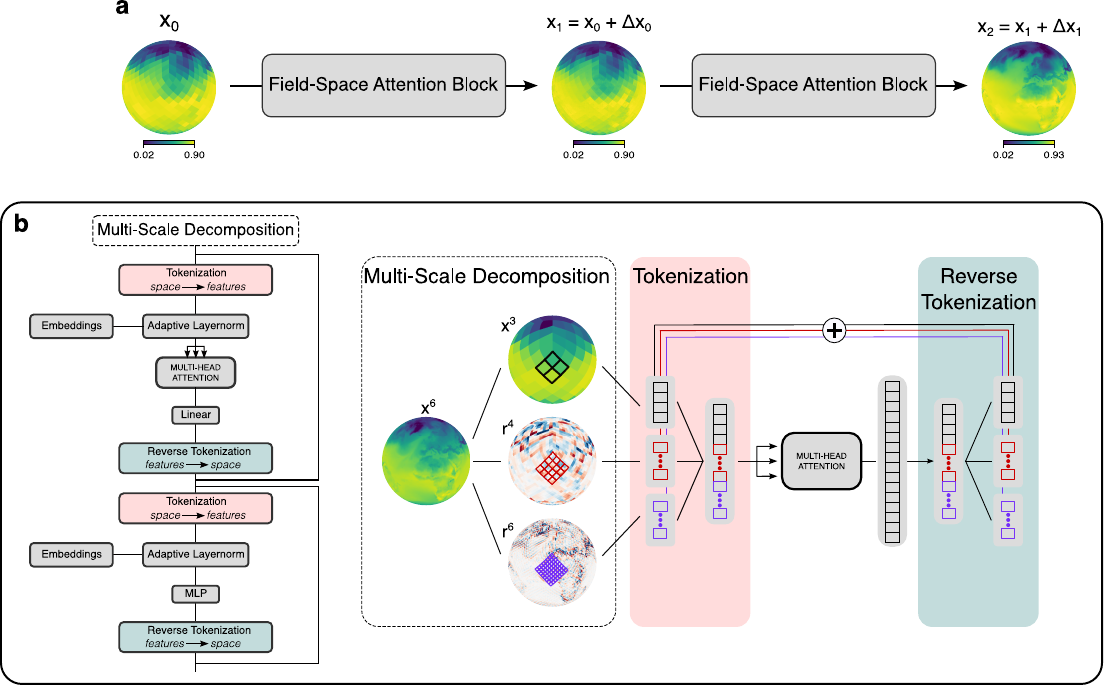}
  \caption{\textbf{Field-Space Attention Block.} 
    \textbf{a}, Exemplary The Field-Space Attention Block is applied to a spherical input field $x_0$ in a super-resolution task. Each block predicts a residual update $\Delta x_n$ that is added to the current state $x_n$ to obtain $x_{n+1} = x_n + \Delta x_n$, gradually refining spatial structures. \textbf{b}, Architecture of a single Field-Space Attention block. The field is first tokenized from space to feature tokens, followed by adaptive layer normalization, multi-head attention, and a linear projection. The updated tokens are then mapped back to space by reverse tokenization. A second residual sublayer repeats the pattern using a MLP instead of attention. The proposed multi-scale attention uses a multi-scale tokenization on the HEALPix sphere, followed by multi-head attention, and updates to each scale by inverse tokenization.}
  \label{fig:field_space_attention}
\end{figure}

We introduce the \emph{Field-Space Attention}, a spatial attention that operates in Field-Space rather than in an abstract latent space (Figure \ref{fig:models}). A Field-Space Attention block applies a deformation to the geophysical input field $x$ toward the target $y$ by:
\begin{equation*}
    x_{n+1} = x_n + \Delta_\theta(x_n),
\end{equation*}
 where $\Delta_\theta(\cdot)$ is produced by our Field-Space Attention block. Consequently, the input field's structure, $x$, is preserved in both the spatial and latent domains, enabling meaningful layer-wise outputs (Figure \ref{fig:field_space_attention} a). \\
Inspired by language models \cite{vaswani2017}, the tokenization in vision tasks has been adopted by learning a latent representation of small patches in Vision Transformers \cite{dosovitskiy2021}. In our Field-Space approach every attention layer uses its own tokenization (Figure \ref{fig:field_space_attention}). We create the tokens by a non-learned multi-scale representation of local patches, which we implemented for the HEALPix sphere \cite{Gorski2004HEALPix}.\\
 On the HEALPix sphere, spatial resolution is organized into a hierarchy of zoom levels $z$, where each pixel at a coarser level is subdivided into four child pixels at the next finer level. Starting from the finest grid, we iteratively average over groups of pixels to obtain coarser parent values, and define the residual at each level as the difference between the fine field and the  coarse field. Repeating this procedure yields a multi-scale decomposition in which the coarsest level captures the large-scale offset, while finer levels encode progressively smaller-scale corrections. Any coarse scale that is defined in the multi-scale decomposition carries $4^{\Delta z}$ additional data points, where $\Delta z=z_\text{in}-z$ defines the difference between the input zoom level $z_\text{in}$ and the coarse scale zoom level $z$. The original field can be reconstructed exactly as the sum of the coarsest field and all residuals. \\
The structure-preserving design of the model allows us to enforce locally zero-mean residuals by subtracting their local mean and transferring this mean to the next coarser grid at any point in the network. This provides a mechanism for conserving the total quantity across scales independent of the task. Upon tokenization, the multi-scale set is stacked a shared grid such that each token carries coarse scale offsets and fine-scale residuals. \\
Exposing all zoom levels as separate but aligned fields gives the multi-head attention direct access to both coarse and fine structure, enabling attention to couple large-scale context with small-scale variability in a geometrically consistent way. 
Our Field-Space Attention block follows the standard Transformer pattern of an attention branch and a feed-forward multi-layer perceptron (MLP) branch in a pre-layer normalization configuration \cite{Xiong2020} (Figure~\ref{fig:field_space_attention}b).
After tokenization embeddings are added to the normalized multi-scale patches by feature-wise linear modulation \cite{Huang2017,Xiong2020}. For creating query, key and values, the multi-scale patches are projected to a defined dimension by linear layers. The output of the multi-head attention is first projected back to the dimension of the multi-scale patches and then added to the input scales. We repeat the same sequence of operations for the feed-forward multi-layer-perceptron branch of the Field-Space Attention block. \\
Note that in the limiting case of a single scale, our attention block reduces to conventional Vision Transformer with a single layer \cite{dosovitskiy2021,lessig2023}. For computational efficiency, we compute the multi-scale decomposition before feeding it to the network. 

\section*{Field-Space Transformers outperform Vision Transformers with fewer parameters}
\begin{figure}[ht]
  \centering
  \def\svgwidth{0.8\columnwidth}  
  \includegraphics[width=0.7\textwidth]{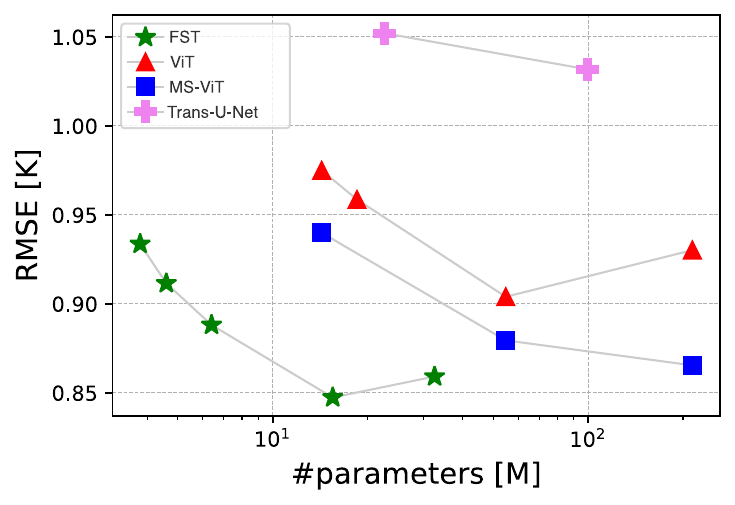}
  \caption{\textbf{Parameter efficiency of Field-Space Transformer compared to baseline architectures.}
Root-mean-square error (RMSE) of super-resolution (x1024) of ERA5 global surface temperatures as a function of the number of trainable parameters for our proposed Field-Space Transformer, a standard Vision Transformer (defined on the HEALPix sphere), a multi-scale ViT which uses our multi-scale tokenization, and a convolutional U-Net with attention layers (Trans-U-Net). Our model consistently achieves lower RMSE at substantially smaller model sizes, indicating superior optimization properties. Note that the largest Field-Space Transformer (32.6 M) overfits to the training dataset, while the largest ViT model exhibits unstable training properties. The numeric values are shown in Supp. Table \ref{tab:comparison}, while a subset of learning curves is shown in Figure \ref{fig:losses_models}.}
  \label{fig:scaling_plot}
\end{figure}

We compare Field-Space Attention to two common architectures that learn in latent-space using attention: the Vision Transformer (ViT), which applies global self-attention to learned patch embeddings, and a (ii) U-Net–based architecture, which uses hierarchical features in global self-attention.\\
In a super-resolution (downscaling) experiment on surface temperatures, we compare architectures as a function of the number of trainable parameters. We used the ECMWF atmospheric reanalysis (ERA5) dataset, downsampling it by a factor of 1024 to create the low-resolution input \cite{Hersbach2020}. We define our Field-Space Transformer with $n_Z=3$ zoom levels, specifically $Z=\{3,5,8\}$ which corresponds to mean cell diameters of $814\,\text{km}$, $204\,\text{km}$ and $ 25\,\text{km}$.\\
Our model outperforms conventional Vision Transformer (ViT) and U-Net–Transformer architectures while using substantially fewer parameters (Figure \ref{fig:scaling_plot}, Table~\ref{tab:comparison_best}) throughout the entire evaluation period (Figure \ref{fig:layer_outputs}). During training, it exhibits rapid convergence and stable dynamics, in contrast to the conventional ViT and U-Net baselines, which show pronounced spikes in the loss curves (Supp. Figure ~\ref{fig:losses_models}).\\
Due to the structure-preserving design of our model, layer-wise outputs can be visualized in a physically meaningful way. Starting from noise-like patterns, the Field-Space Attention gradually adds coherent patterns with increasing amplitude (Supp. Figure .~\ref{fig:layer_outputs}).  Increasing the number of parameters accelerates convergence but also leads to earlier overfitting. With 32.6 million parameters, the model starts to overfit the training data after about 60{,}000 iterations. Larger models exhibit this behavior even earlier. Unlike models that operate in an increasingly high-dimensional latent space, Field-Space Attention benefits from learning richer cross-scale interactions through multi-scale attention.\\
To isolate the effect of multi-scale tokenization, we replace the standard tokenization in the ViT architecture with our multi-scale tokenization (MS-ViT). This stabilizes training and improves accuracy, especially for models with a large number of parameters.\\
In the limit $n_Z=1$, the multi-scale decomposition collapses to a single scale and our Field-Space Transformer reduces to a stack of Vision Transformer layers. As shown in Supp. Table~\ref{tab:comparison_Zooms} and Figure \ref{fig:losses_zooms}, the accuracy of our Field-Space Transformer improves systematically with the number of zoom levels $n_Z$. In the limiting case of stacked Vision Transformers ($n_Z=1$), the model does not converge beyond a relatively high error level, whereas adding zoom levels increases accuracy, with gains that gradually saturate as more levels are added.

\begin{figure}[ht]
  \centering
  \includegraphics[width=0.8\textwidth]{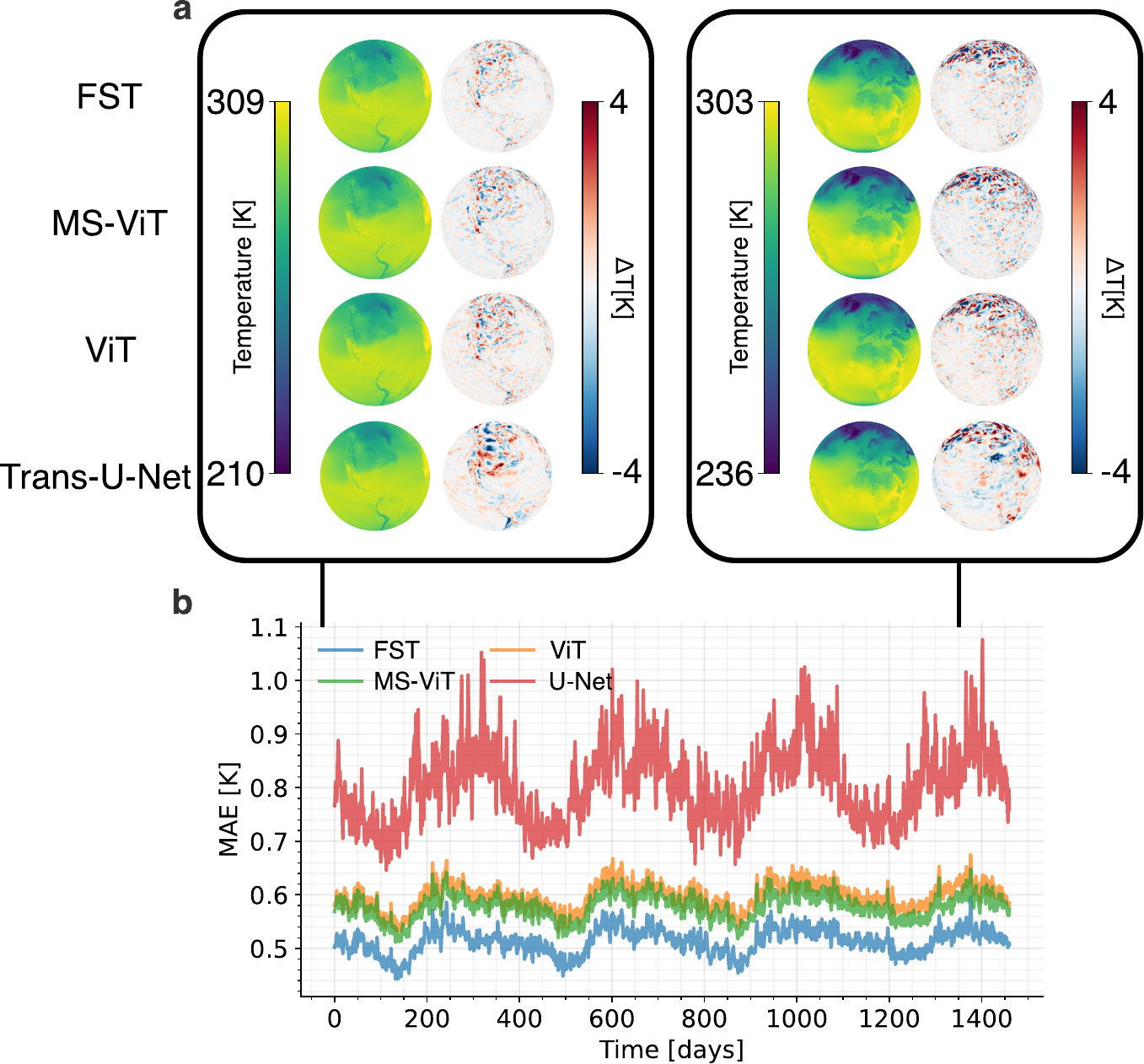}
  \caption{\textbf{Temporal performance and example snapshots of selected models.} 
\textbf{a} Super-resolution snapshots for two days from our Field-Space Transformer (FST, $15.5$\,M ), Multi-Scale Vision Transformer (MS-ViT, $54.8$\,M), Vision Transformer (ViT, $54.8$\,M), and Transformer-U-Net ($99.6$\,M). For each model we show the reconstructed temperature field in the left and the corresponding error field $\Delta T$ in the right column. \textbf{b} Time series of the daily mean absolute error (MAE) over the full evaluation period. The Field-Space Transformer attains the lowest and most stable MAE, followed by MS-ViT and ViT, while U-Net shows substantially larger and more variable errors}
  \label{fig:layer_outputs}
\end{figure}

\section*{Links to Multiscale Numerical Discretizations}
\label{sec:multiscale_connections}

Field-Space Transformer is designed around an explicit multilevel representation of the prognostic field, in which a coarse, cell-averaged component is complemented by a hierarchy of zero-mean residuals. This construction closely parallels conservative finite-volume discretizations and mesh coarsening/refinement procedures, where restriction and prolongation operators are chosen to preserve cell integrals under changes in resolution \cite{Künze2012A}. It is also directly related to scale-separated formulations in large-eddy simulation (LES). The resolved field carries the large-scale dynamics, while subgrid fluctuations are constrained to have zero mean and influence the evolution predominantly through flux-like coupling terms between scales \cite{Gravemeier2006Scale-separating,Hu2011Scale}. In geophysical fluid dynamics, the coarse component can be interpreted as the large-scale background circulation and the residual levels as eddies and localized anomalies. Their interactions correspond to scale-to-scale transfers that are central to atmospheric and oceanic turbulence \cite{Zhang2024Scale-to-scale,Storer2023Global,Aluie2017Mapping,Grooms2018Multiscale}.\\
From an algorithmic viewpoint, the same principle underlies multigrid and multilevel solvers for partial differential equations, which distribute solution components and residuals across a hierarchy of grids and iteratively exchange information between coarse and fine levels \cite{Napolitano1986Incremental, Abu-Labdeh2022Monolithic}.  Closely related decompositions appear in image and signal processing, for example in Laplacian and wavelet pyramids that represent a smooth base signal together with scale-localized detail coefficients \cite{Paris2011Local,Wahyuni2015Wavelet}.\\
These connections highlight the key distinction from latent-space multi-resolution token hierarchies. In Field-Space Transformer, the hierarchy is fixed by a physically meaningful restriction/refinement scheme, and all transformations act directly on this multiscale field representation. Intermediate states therefore remain interpretable as fields on the sphere, and cross-scale attention can be viewed as learning data-driven inter-level couplings analogous to modeled inter-scale exchanges in classical multiscale methods. More broadly, the approach matches standard geophysical practice, where dynamics are expressed as compositions of physical-space operations (filtering, restriction, reconstruction, and inter-scale coupling) rather than as trajectories in an abstract latent space.

\begin{table}
\caption{Root-mean-squared error of our Field-Space Transformer (FST) in comparison to the best performing Vision Transformer(ViT), Multi-Scale Vision Transformer (MS-ViT), and U-Net Transformer. Our 32.6M Field-Space Transformer  model was already overfitting. The result of all models is shown in Supp. Table \ref{tab:comparison}.}
\label{tab:comparison_best}
\begin{tabular}{lr|r|r|rrr}
\toprule
family & \multicolumn{1}{c}{ViT} & \multicolumn{1}{c}{MS-ViT}& \multicolumn{1}{c}{U-Net} & \multicolumn{3}{c}{FST (ours)}   \\
\# params [M] & 54.8 & 214.0 & 99.6  & 6.4 & 15.5 & 32.6 \\
\midrule
RMSE [K] & 0.904 & 0.865 & 1.032 & 0.888 & \textbf{0.847} & 0.859* \\
\bottomrule
\end{tabular}
\end{table}

\section*{Discussion}
The Field-Space Transformer differs from the standard latent-space approach in that it operates directly on the input field via a fixed, multiscale decomposition. Instead of learning a sequence of transformations on abstract tokens, the model learns deformations that preserve the structure of a multilevel field representation on the sphere.\\
Our experiments show that the Field-Space Attention is outperforming strong Vision Transformer and Transformer-U-Net baselines in terms of accuracy, while having substantially fewer parameters. Early layers of the Field-Space Transformer inject small-scale, noise-like perturbations around the coarse field, which are progressively shaped into coherent fine-scale structures (Figure~\ref{fig:layer_outputs}). This iterative refinement process is similar to that used in denoising diffusion models, which transform noisy states into realistic samples through repeated, incremental updates \cite{ho2020denoising}, suggesting a natural link to diffusion-style generative formulations on the sphere.\\
The key mechanism in Field-Space Transformer is the non-learned, joint multi-scale tokenization across resolutions. Existing multi-scale Vision Transformer architectures often build hierarchical latent tokens that are processed at different resolutions but are not explicitly tied to a fixed physical decomposition of the input \cite{liu2021swin,Li2024,carlsson2024}. In contrast, Field-Space Transformer uses a fixed multi-scale decomposition to define tokens at several zoom levels that are strictly aligned on a shared grid, and all attention and multi-layer perceptron operations act on this joint representation. This design has two main consequences. First, it stabilizes optimization. Replacing standard patch embeddings with our multi-scale tokenization improves convergence speed and accuracy in a conventional Vision Transformers, and Field-Space Transformer becomes more stable and accurate as the number of scales increases. In the single-scale limit ($n_Z = 1$), the architecture reduces to a stacked ViT-like model that, in our experiments, did not convergence. Second, joint multi-scale tokenization enables coherent cross-scale information flow. Each token attends simultaneously to coarser and finer scales, allowing the model to combine large-scale context with local variability in a geometrically consistent way. This property is particularly important for climate-related tasks, where local processes are often strongly influenced by global-scale dynamics.\\
The Field-Space Transformer architecture also supports controlled trade-offs between the number of parameters and the number of patches. This trade-off is governed by the tokenization zoom level $z_A$ and by the set of zoom levels used in the Field-Space Attention blocks. In this work, we use the same set of zoom levels for queries, keys and values across all layers. Under this configuration, the number of parameters in a Field-Space Attention block grows approximately linearly with the number of pixels per patch. At very high spatial resolution this leads to parameter growth dominated by the dense projection layers acting on the multi-scale tokens. The non-learned decomposition provides a way to integrate optimisation strategies that reduce memory and parameter footprint. One strategy is to use smaller patch sizes at higher zoom levels, combined with local or shifted-window attention, similar to Swin Transformers \cite{liu2021swin}. A second strategy is to compress high-resolution residuals before they enter attention blocks. Because the decomposition enforces zero-mean residuals around coarser fields, these residuals contain less redundant information and are promising targets for learned compression. Appropriate choices of patch size, attention locality and residual compression could make Field-Space Transformer even more parameter-efficient and extend its applicability to extremely high resolutions, approaching kilometer-scale grids.
\begin{table}[!h]
\caption{Root-mean-squared error of our 3.8M Field-Space Transformer with a different number of zoom levels ($n_Z$). The corresponding training and validation losses are shown in Supp. Figure \ref{fig:losses_zooms}.}
\captionsetup{width=0.7\linewidth}
\label{tab:comparison_Zooms}
\begin{tabular}{c|c|c|r}
\toprule
$n_Z$ & $Z$ & \# params [M] & RMSE [K] \\
1 &  $8$ & 3.8 & 1.653\\
2 &  $3,8$ & 3.8 & 0.945\\
3 &  $3,4,8$ & 3.8 & 0.934\\
4 &  $3,4,5,8$ & 3.8 & 0.913	\\
5 &  $3,4,5,6,8$ & 4.1 & 0.908 \\
\bottomrule
\end{tabular}
\end{table}\\ The explicit scale decomposition provides a natural scale-aware inductive bias. By enforcing that each zoom level carries only residual information and that mean components are propagated to coarser levels, cross-scale consistency in a task-agnostic way is enforced. This is related to hard constraints to improve generalization in super-resolution, where the low-resolution input is enforced as an exact average constraint on the output \cite{Harder2023}. In our case, the constraint is built directly into the representation and does not depend on a specific input–output pairing. The same mechanism can therefore be reused across tasks such as prediction, infilling and data assimilation. More generally, imposing scale-local constraints on internal representations rather than only on final outputs offers a flexible way to encode physical priors about conservation, balance and smoothness at different scales. \\
These properties make Field-Space Transformer a promising building block for foundation models of continuous fields. The multi-scale decomposition supports learning relations between multiple variables and modalities stacked along the channel dimension via cross-variable attention \cite{lessig2023}. The same framework extends to four-dimensional space–time data, where different zoom levels could carry different temporal contexts. For autoregressive prediction, coarse levels could provide long time horizons at low spatial resolution, while fine levels resolve short, detailed context windows. We expect this to improve stability and physical coherence in longterm predictions. Because the decomposition is non-learned, it is also possible, in principle, to introduce higher spatial resolutions after pretraining, without retraining the full model.\\
Our empirical evaluation focuses on a single super-resolution task. The underlying architectural principles are, however, not specific to this setting. Any task involving continuous fields on regular or spherical grids—for example precipitation, wind, ocean currents, aerosol concentrations or non-geophysical scalar fields—can be represented within the same multi-scale, structure-preserving framework. A similar conversion of pixels into multi-scale spatial tokens, combined with learning deformations in image space rather than in purely latent space, may also benefit conventional high-resolution image tasks. Future work will therefore extend evaluation to broader applications, including spatiotemporal prediction, data assimilation and diffusion-based generative modeling under stronger physical constraints and multi-variable coupling.\\
In summary, Field-Space Transformer provides a field-centric alternative to latent-space abstraction for modeling continuous geophysical fields. By learning structure-preserving deformations on a fixed multi-scale spherical decomposition, the architecture achieves strong accuracy and favorable optimization behavior, while maintaining interpretability and alignment with physical structure. Its scalability, support for task-agnostic scale constraints and compatibility with multi-variable and space–time settings suggest that the Field-Space Transformer could be a key component in future Earth system foundation models, as well as in any other vision task involving continuous data.

\section*{Methods}

\subsection*{HEALPix Grid}

The \emph{HEALPix} (\emph{Hierarchical Equal Area isoLatitude Pixelisation}) grid is a spherical discretization scheme that divides the sphere into equal-area cells, or pixels.  
It is widely used in astrophysics for representing all-sky data at multiple resolutions while preserving uniform area coverage and iso-latitude structure \cite{Gorski2004HEALPix}.

A HEALPix map is parameterized by a resolution parameter \( N_{\text{side}} \), which must be a power of two.  
The total number of pixels on the sphere is given by
\[
N_{\text{pix}} = 12\,N_{\text{side}}^2.
\]
To describe resolution hierarchically, we use the discrete \emph{zoom level} \( z \in \mathbb{Z}_{\ge 0} \).  
Each increment in zoom level doubles the linear resolution on each HEALPix face, and therefore quadruples the number of pixels across the full sphere:
\[
N_{\text{side}}(z) = 2^z,
\qquad
N_{\text{pix}}(z) = 12 \cdot 4^z.
\]
Each pixel at level \(z\) is thus subdivided into four child pixels at level \(z+1\).  
This nested structure defines a natural hierarchical tree, enabling efficient aggregation or refinement of data across multiple spatial scales on the sphere.
We denote by \(z_{\min}\) and \(z_{\text{in}}\) the coarsest and finest zoom levels considered, respectively, and define the total number of hierarchical scales as
\[
L = z_{\text{in}} - z_{\min} + 1.
\]

\subsection*{Multi-Scale Decomposition}

Let \(x^{(z_{\text{in}})} \in \mathbb{R}^{B \times N_{z_{\text{in}}} \times C}\) denote a signal or feature map defined on the HEALPix grid at the finest zoom level \(z_{\text{in}}\), with batch size \(B\) and channel dimension \(C\).  
We decompose this signal into a sequence of coarser representations \(\{x^{(z)}\}_{z=z_{\min}}^{z_{\text{in}}}\) and corresponding residuals \(\{r^{(z)}\}\), which together form a hierarchical, multi-scale representation.

At each step from a finer level \(z_{\text{in}}\) to a coarser level \(z\), the grid is partitioned such that each coarse pixel aggregates \(4^{(z_{\text{in}} - z)}\) fine pixels.  
The coarser representation is computed as the mean over these groups:
\[
x^{(z)}_p = \frac{1}{4^{(z_{\text{in}} - z)}} 
\sum_{q \in \text{children}(p)} x^{(z_{\text{in}})}_q,
\]
where \(\text{children}(p)\) denotes the set of finer-level pixels that map to the coarse pixel \(p\).  
The corresponding residual captures the high-frequency content removed by averaging:
\[
r^{(z)}_q = x^{(z_{\text{in}})}_q - x^{(z)}_{\lfloor q / 4^{(z_{\text{in}} - z)} \rfloor}.
\]

Algorithmically, this decomposition proceeds iteratively from fine to coarse levels:
\begin{enumerate}
  \item Group the fine-scale pixels corresponding to the next coarser level \(z\).
  \item Compute the mean within each group to obtain \(x^{(z)}\).
  \item Subtract the mean from the fine representation to obtain residuals \(r^{(z)}\).
\end{enumerate}

The process yields a hierarchical set of tensors
\[
\Bigl\{\, x^{(z_{\min})}, \; \{\, r^{(z)} \mid z_{\min} < z \le z_{\text{in}} \,\} \Bigr\},
\]
where \(x^{(z_{\min})}\) represents the coarsest-scale average, and each \(r^{(z)}\) encodes the residual (detail) information specific to zoom level \(z\).

The original signal at the finest resolution can then be reconstructed as the sum of upsampled contributions from all scales:
\[
x^{(z_{\text{in}})} 
= 
\mathcal{U}_{z_{\min} \to z_{\text{in}}}\!\bigl(x^{(z_{\min})}\bigr)
\;+\;
\sum_{z = z_{\min} + 1}^{z_{\text{in}}}
\mathcal{U}_{z \to z_{\text{in}}}\!\bigl(r^{(z)}\bigr),
\]
where \(\mathcal{U}_{z \to z_{\text{in}}}\) denotes an upsampling operator that expands features from zoom \(z\) to the finest grid by repeating each value \(4^{(z_{\text{in}} - z)}\) times along the pixel axis.

This hierarchical decomposition provides a compact, interpretable representation of spherical data across \(L\) zoom levels, analogous to a Laplacian or wavelet pyramid on the HEALPix sphere \cite{Wahyuni2015Wavelet}.

\subsection*{Scale Conservation}

After the multi-scale decomposition, we apply a \emph{scale constraining} operation (SC) that enforces consistency between zoom levels.  
This operation ensures that fine-scale residuals have zero mean within each parent region at any coarser zoom \(z' < z\), while the corresponding mean values are transferred to that coarser level.  
As a result, large-scale (low-frequency) information accumulates at coarser zooms, and small-scale (high-frequency) variations remain locally zero-mean.

Let \(x^{(z)} \in \mathbb{R}^{B \times N_{z} \times C}\) denote the feature map at zoom level \(z\).  
For any pair of levels \((z, z')\) with \(z' < z\), each coarse pixel \(p\) at level \(z'\) corresponds to a group of 
\[
|\text{children}(p)| = 4^{(z - z')}
\]
fine-scale pixels at level \(z\), denoted by \(\text{children}_{z \to z'}(p)\).  
The mean of these fine-scale pixels is defined as
\[
m^{(z \to z')}_p 
= 
\frac{1}{4^{(z - z')}} 
\sum_{q \in \text{children}_{z \to z'}(p)} x^{(z)}_q.
\]

The fine-scale features are then recentered by subtracting their local mean:
\[
\tilde{x}^{(z)}_q = x^{(z)}_q - m^{(z \to z')}_{\lfloor q / 4^{(z - z')} \rfloor},
\]
where \(q\) indexes the fine-scale pixels at level \(z\), and \(\lfloor q / 4^{(z - z')} \rfloor\) denotes the index of their parent pixel at level \(z'\).

The removed mean values are added to the corresponding coarse pixels:
\[
\tilde{x}^{(z')}_p = x^{(z')}_p + m^{(z \to z')}_p.
\]

This guarantees that the mean of all fine-scale residuals under each parent pixel is zero:
\[
\frac{1}{4^{(z - z')}} \sum_{q \in \text{children}_{z \to z'}(p)} \tilde{x}^{(z)}_q = 0,
\]
while conserving the total feature sum across the two scales:
\[
\sum_{q \in \text{children}_{z \to z'}(p)} x^{(z)}_q 
+ 4^{(z - z')} x^{(z')}_p
=
\sum_{q \in \text{children}_{z \to z'}(p)} \tilde{x}^{(z)}_q
+ 4^{(z - z')} \tilde{x}^{(z')}_p.
\]

In practice, this operation can be applied iteratively across an arbitrary sequence of zoom levels,
\[
z_{\text{in}} \to z_{\text{in}} - k_1 \to z_{\text{in}} - k_2 \to \dots \to z_{\min},
\]
where the step sizes \(k_i \ge 1\) need not be uniform.  
At each stage, the operation enforces zero-mean fine-scale residuals relative to the chosen coarser level, ensuring that mean information is pushed upward while maintaining global conservation.

The resulting hierarchy is \emph{scale-conservative}:  
each finer level encodes locally zero-mean information, while coarse levels store progressively aggregated mean components.

\subsection*{Field-Space Attention with Multi-Scale Tokenization}

Starting from the scale-constrained representations  
\(\{\, \tilde{x}^{(z)} \mid z_{\min} \le z \le z_{\text{in}} \,\}\),  
we introduce the \emph{Field-Space Attention} mechanism, which enables joint processing of spherical features across multiple zoom levels on a shared token grid.

We define an \emph{attention zoom level} \(z_A\), which determines the spatial resolution and the number of tokens \(N_{z_A}\) on which attention is computed.  
For any finer level \(z \ge z_A\), a reshaping operator \(\mathcal{R}_{z \to z_A}\) maps the feature map \(\tilde{x}^{(z)}\) onto the token grid of level \(z_A\) by grouping the \(4^{(z - z_A)}\) child pixels belonging to each parent pixel into the feature dimension:
\[
\mathcal{R}_{z \to z_A}\!\bigl(\tilde{x}^{(z)}\bigr)
\in \mathbb{R}^{B \times N_{z_A} \times 4^{(z-z_A)}}.
\]
Since each input map has a single feature channel, this operation directly stacks spatially corresponding values without increasing the channel dimension.

To construct the input for the attention mechanism, we define two (possibly different) subsets of zoom levels,
\(\mathcal{Z}_q \subseteq \{z_A, \dots, z_{\text{in}}\}\)
and
\(\mathcal{Z}_{kv} \subseteq \{z_A, \dots, z_{\text{in}}\}\),
which determine the ranges of scales used for the query and key–value pathways, respectively.  
All selected feature maps are reshaped onto the \(z_A\) grid and concatenated along the feature axis:
\[
X_q^{(z_A)} =
\operatorname{Concat}\Bigl(\{\mathcal{R}_{z \to z_A}(\tilde{x}^{(z)}) \mid z \in \mathcal{Z}_q\}\Bigr)
\in \mathbb{R}^{B \times N_{z_A} \times C_q},
\]
\[
X_{kv}^{(z_A)} =
\operatorname{Concat}\Bigl(\{\mathcal{R}_{z \to z_A}(\tilde{x}^{(z)}) \mid z \in \mathcal{Z}_{kv}\}\Bigr)
\in \mathbb{R}^{B \times N_{z_A} \times C_{kv}},
\]
with total feature dimensions
\[
C_q = \sum_{z \in \mathcal{Z}_q} 4^{z - z_A}
\]

\[
C_{kv} = \sum_{z \in \mathcal{Z}_{kv}} 4^{z - z_A}
\]
In Figure \ref{fig:field_space_attention}c, $z_A=2$ with $Z=\{3,4,6\}$ results in $C=4^1+4^2+4^4=276$, compared to $4^4=256$ in case of a stack of conventional vision transformers with $Z=\{6\}$.
Linear projections then map the concatenated tensors to the query, key, and value embeddings:
\[
Q = W_Q X_q^{(z_A)}, 
\qquad
K = W_K X_{kv}^{(z_A)}, 
\qquad
V = W_V X_{kv}^{(z_A)}.
\]
 Multi-head attention is computed across the \(z_A\)-level tokens as
\[
\mathrm{CSA}(X_q^{(z_A)}, X_{kv}^{(z_A)})
=
\operatorname{Softmax}\!\left(\frac{QK^\top}{\sqrt{d}}\right) V,
\]
where \(d\) denotes the per-head embedding dimension.  
The output is linearly projected and added back to the query pathway through a residual connection:
\[
Y^{(z_A)} = X_q^{(z_A)} + W_O \, \mathrm{CSA}(X_q^{(z_A)}, X_{kv}^{(z_A)}).
\]

Intuitively, this formulation allows each token at the chosen attention level to jointly access information from both coarser and finer scales, depending on the selected subsets \(\mathcal{Z}_q\) and \(\mathcal{Z}_{kv}\).  
By aligning all scales on a shared grid, CSA achieves efficient and spatially consistent interaction between hierarchical representations, enabling the model to capture local fine-scale structure while remaining sensitive to large-scale contextual patterns across the sphere.


\subsection*{Position Embeddings}
As position embeddings, we use learnable per-grid-point embeddings similar to Brenowitz et al. \cite{BrenowitzEtAl2025}. Instead of random initializaion, we chose to initialize the learnable embeddings using \textit{spherical harmonics}. 
For each feature channel, a random pair of spherical harmonic indices $(l, m)$ is sampled, where $l$ determines the spatial frequency 
and $m$ the azimuthal order. The corresponding real-valued harmonic basis function $Y_l^m(\theta, \phi)$ is then evaluated 
at the grid coordinates to generate the embedding values:
\[
Y_l^m(\theta, \phi) = 
\begin{cases}
\sqrt{2}\,\mathrm{Re}\{Y_l^{|m|}(\theta, \phi)\}, & m > 0, \\[6pt]
Y_l^0(\theta, \phi), & m = 0, \\[6pt]
\sqrt{2}\,\mathrm{Im}\{Y_l^{|m|}(\theta, \phi)\}, & m < 0.
\end{cases}
\]

The spectral range of spherical harmonics is determined by the maximum angular frequency $l_{\max}$, which depends on the smallest resolvable angular scale of the HEALPix grid:
\[
d\theta = \sqrt{\frac{4\pi}{n_\text{cells}}}\,=\sqrt{\frac{4\pi}{12\times4^z}},
\]
where $n_\text{cells}$ is the total number of grid cells on the sphere and $z$ the zoom level.

The corresponding Nyquist limit in spectral space is then computed as
\[
L_{\mathrm{nyq}} = \left\lfloor \frac{\pi}{2\,d\theta} \right\rfloor.
\]
To ensure that at least one nontrivial harmonic degree is sampled, we choose:
\[
l_{\max} = \max(1, L_{\mathrm{nyq}}).
\]
For each sampled degree $l$, the azimuthal index $m$ is uniformly sampled from the discrete interval $[-l,\,l]$. \\
Positional embeddings are added after adaptive layer norm (AdaLN) using feature-wise linear modulation (FiLM) \cite{Huang2017,Perez2017}. Since large fields can have a lot of features, we restrict the dimension of per-grid-point learned feature vectors to a defined dimension, e.g. $64$, and project it to the dimension of the hierarchical patches $C_q$ and $C_kv$.

\subsection*{Experiments}

\paragraph{Data and Setup}
We perform downscaling experiments on ERA5 reanalysis \cite{Hersbach2020} data, originally provided at a spatial resolution of $1280 \times 640$. For the ViT-based models, we remap the data onto the HEALPix grid at zoom level $z = 8$. We downsample both data sources by a factor of $\times 1024$. All models are trained for {100k iterations} on {4 NVIDIA A100 GPUs (80~GB)} on the period 1941-2021. We keep 2021-04/2025 for evaluation. Training is performed using {mixed precision} (\texttt{bf16-mixed}) for efficiency, while validation is always conducted in {full precision} to ensure numerical consistency. We use a basic \textit{mean-squared error} loss for the training of all models. 

\paragraph{Field-Space Transformer}
We are testing different configuration of our Field-Space Transformer, listed in Table \ref{tab:model_configs}. For analyzing the influence of the multi-zoom decomposition we define the model on different zoom levels $n_Z$. For the models ability to scale with the number of learned parameters, we chose $n_Z=3$ with $Z=\{3, 5, 8\}$, with different model dimensions $d$ in query, key and value projections. For tokenization we chose $z_A = 3$.

\begin{table}[h]
\caption{Model configurations tested. The $n_Z=3$ configuration is used for analyzing the models scaling ability.}
\captionsetup{width=0.7\linewidth}
\label{tab:model_configs}
\begin{tabular}{l |c |c}
\toprule
 $n_Z$ & $Z$ & model dimension $d$ (\# layers) \\[2mm] 
3 &3,5,8 &  32 (5), 32 (7), 64 (7), 128 (7)\\  
& & \\
1 & 3 & 32 (5) \\
2 & 3,8 & 32 (5) \\
4 & 3,4,5,8 & 32 (5) \\
5 & 3,4,5,6,8 & 32 (5)\\
\bottomrule
\end{tabular}
\end{table}
For simplicity, we choose a constant $Z$ and $z_A$ across all query, key, and value projections. We use a learning rate of $2\cdot 10^{-4}$ and a cosine learning rate scheduler with a warmup of $2000$ iterations and a batch-size of $16$.

\paragraph{Vision Transformer (ViT)}
We define the baseline Vision Transformer using a local patch-size of $4^{5}=1024$, resulting in tokens on zoom level $z = 3$, which corresponds to the token grid of our Field-Space Transformer. This is the closest value to Lessig et al. \cite{lessig2023} who used a input token size of $729$ for temperature. We use the same positional embedding like the Field-Space Transformer. This architecture follows the canonical ViT design but is adapted to operate directly on the HEALPix grid. We tested learning rates of $2\cdot 10^{-4}$ and $2\cdot 10^{-5}$ and a cosine learning rate scheduler with a warmup of $2000$ iterations and a batch-size of $16$. 

\paragraph{Multi-Scale Vision Transformer (MS-ViT)}
To isolate the contribution of the {scale decomposition}, we integrate the proposed multi-scale tokenization into the standard Vision Transformer architecture. The decomposition is applied both at the {input} (token embedding) and output stages. We used a learning rates of $2\cdot 10^{-4}$ and a cosine learning rate scheduler with a warmup of $2000$ iterations and a batch-size of $16$. 

\paragraph{Transformer U-Net}
For comparison, we employ a four and a five layer {Transformer U-Net} that combines {ResNet-style convolutional blocks \cite{szegedy2017inception}} with {spatial attention} at the deepest level. The model is optimized using the Adam optimizer with a learning rate of $5 \times 10^{-4}$. For evaluation, we use a latitude-weighted RMSE.

\section*{Acknowledgments}
This project received funding from the German Federal Ministry of Research, Technology and Space (BMBFTR) under grant 16ME0679K. Supported by the European Union–NextGenerationEU; and in part by the Horizon Europe project EXPECT under Grant No. 101137656. This work used resources of the Deutsches Klimarechenzentrum (DKRZ) granted by its Scientific Steering Committee (WLA) under project ID bk1372.


\bibliography{sn-bibliography}

\section*{Supplementary Material}
\begin{appendices}

\begin{figure}[ht]
  \centering
  \includegraphics[width=0.8\textwidth]{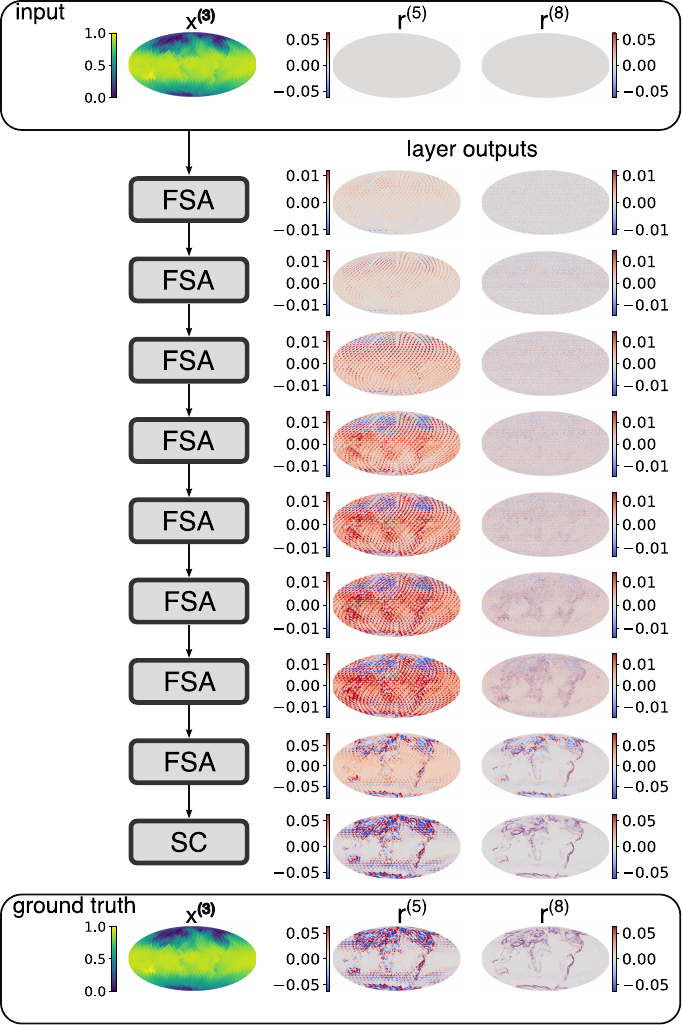}
  \caption{\textbf{Layer-wise residual fields of our Field-Space Transformer.} 
Layer outputs of the $15.5$\,M parameter Field-Space Transformer with zoom levels $Z=\{3,5,8\}$ for temperature super-resolution by a factor of $1024$. The panels show the residual fields $r^{(5)}$ and $r^{(8)}$ (zoom levels 5 and 8) across the stack of Field-Space Attention blocks (FSA) and the final scale-constraining layer (SC; bottom). Zoom level 3 is omitted for clarity, since the data are defined on this grid and the learned updates at this scale remain close to zero ($\approx 10^{-8}$). The model gradually builds up structured corrections at the finer scales that match the ground-truth residual fields (bottom row).}
  \label{fig:layer_outputs}
\end{figure}

\begin{figure}[ht]
  \centering
  \includegraphics[width=1.0\textwidth]{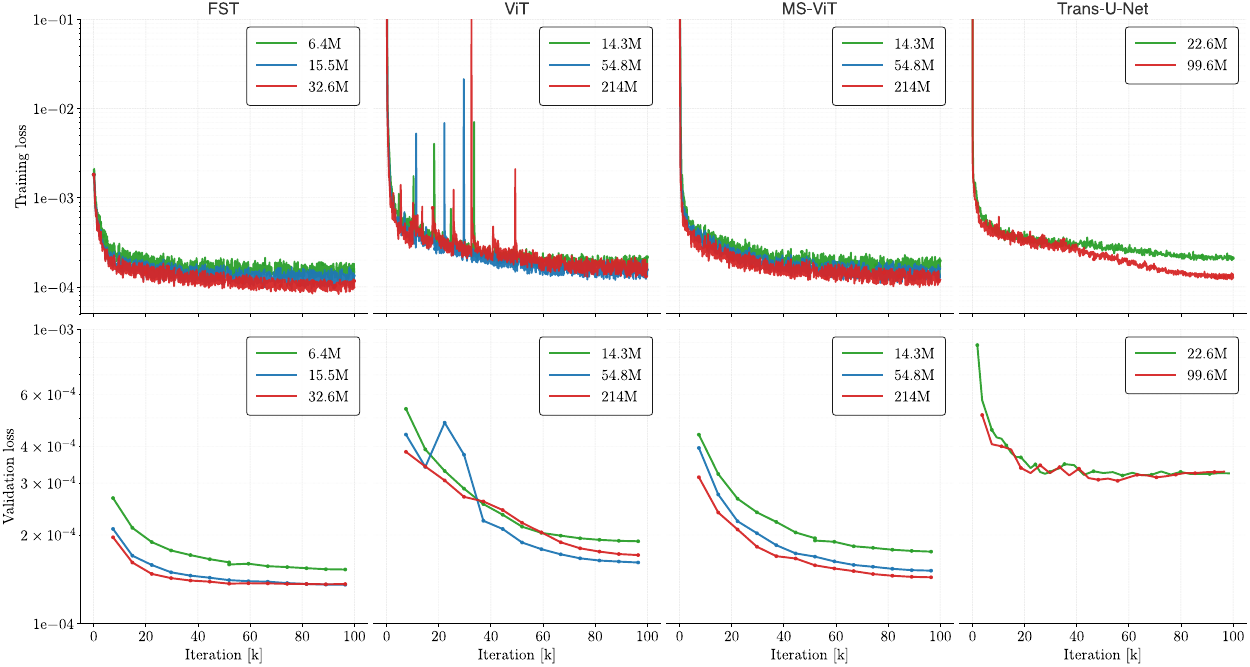} 
  \caption{\textbf{Optimization behaviour across architectures and model sizes.} 
Training (top row) and validation (bottom row) losses for the Field-Space Transformer (FST), a baseline Vision Transformer (ViT), a Vision Transformer with multi-scale tokenization (MS-ViT), and a U-Net with residual connections and Transformer blocks, each trained on the temperature super-resolution task with a different number of parameters. Field-Space Transformer shows fast and smooth convergence across all sizes, while the single-scale ViT exhibits pronounced training instabilities with large loss spikes. MS-ViT improves optimization over ViT but requires more parameters than Field-Space Transformer, and U-Net converges more slowly and to higher validation loss.}
  \label{fig:losses_models}
\end{figure}

\begin{figure}[ht]
  \centering 
  \includegraphics[width=0.8\textwidth]{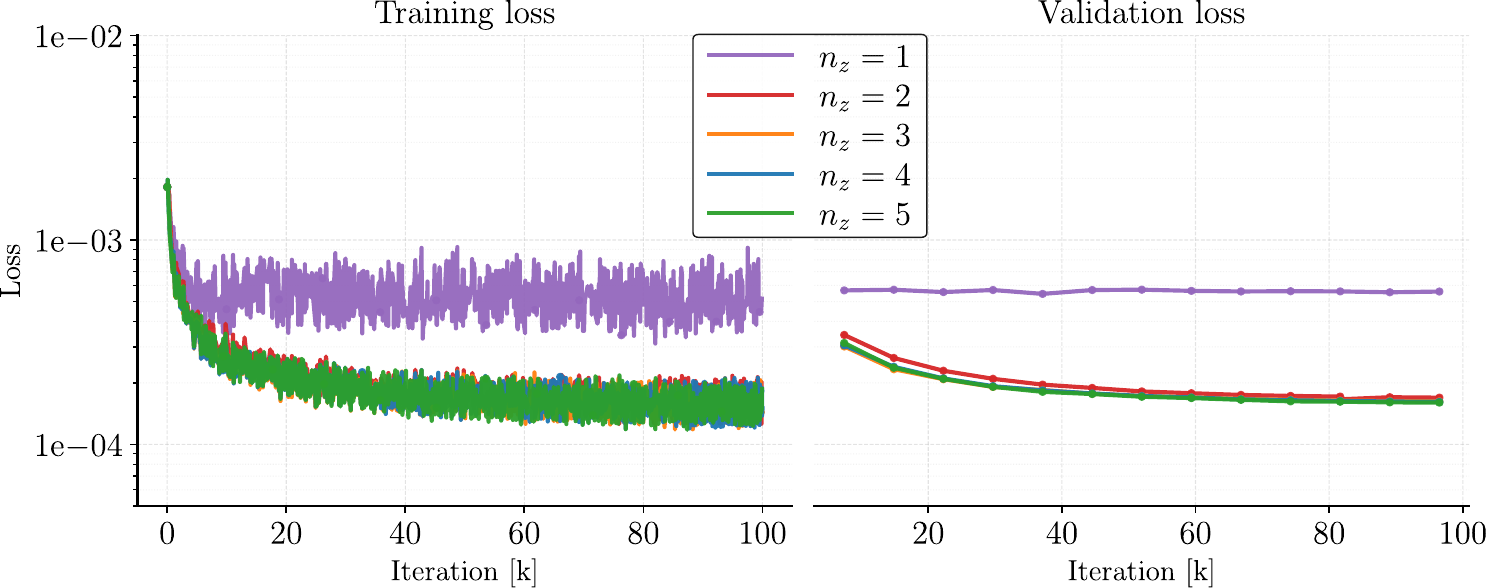}
  \caption{\textbf{Effect of the number of zoom levels on optimization.} 
Training (left) and validation (right) losses of the Field-Space Transformer for different numbers of zoom levels $n_Z$ (see Table~\ref{tab:model_configs}). The configuration with $n_Z = 3$ corresponds to the $3.8$M parameter model in Table~\ref{tab:comparison_best}. All multi-scale variants with $n_Z \geq 2$ show stable optimization and similar final validation loss, while the single-scale model ($n_Z = 1$) converges to a substantially higher loss. The corresponding test-set RMSE values are reported in Table~\ref{tab:comparison_Zooms}.}
  \label{fig:losses_zooms}
\end{figure}

\begin{table}[ht]
\centering
\caption{Root-mean-squared error of our Field-Space Transformer (FST) model in comparison to a Vision Transformer (ViT), Multi-Scale Vision Transformer (MS-ViT), and U-Net Transformer.}
\label{tab:comparison}

\footnotesize
\setlength{\tabcolsep}{2.5pt}      
\renewcommand{\arraystretch}{0.95} 

\begin{tabular}{@{}lrrrr|rrr|rrrrr|rr@{}}
\toprule
family & \multicolumn{4}{c}{ViT} & \multicolumn{3}{c}{MS-ViT} & \multicolumn{5}{c}{\textbf{FST (ours)}} & \multicolumn{2}{c}{U-Net} \\
\# params [M] & 14.3 & 18.5 & 54.8 & 214.0 & 14.3 & 54.8 & 214.0 & 3.8 & 4.6 & 6.4 & 15.5 & 32.6 & 22.6 & 99.6 \\
\midrule
RMSE & 0.993 & 0.959 & 0.904 & 0.930 & 0.975 & 0.879 & 0.865 & 0.934 & 0.912 & 0.888 & 0.847 & 0.859 & 1.052 & 1.032 \\
bias & 0.047 & 0.006 & 0.003 & 0.018 & -0.026 & -0.015 & 0.003 & -0.004 & 0.006 & 0.002 & 0.000 & 0.002 & 0.086 & 0.126 \\
\bottomrule
\end{tabular}
\end{table}




\end{appendices}

\end{document}